\documentclass{article}

\usepackage{arxiv}
\usepackage[utf8]{inputenc} 
\usepackage[T1]{fontenc}    
\usepackage{hyperref}       
\usepackage{url}            
\usepackage{booktabs}       
\usepackage{amsfonts}       
\usepackage{nicefrac}       
\usepackage{microtype}      
\usepackage{lipsum}		
\usepackage{graphicx}
\usepackage{doi}
\usepackage{amsmath}
\usepackage{algorithm}
\usepackage{xcolor}

\usepackage{graphicx}
\usepackage{scalerel, amsmath}
\usepackage{amssymb}
\usepackage{booktabs}
\usepackage{verbatim}
\usepackage{pifont}
\usepackage{dsfont}
\usepackage{float}
\usepackage[english]{babel}
\usepackage{amsthm}
\usepackage{xcolor}
\usepackage{multicol}
\usepackage{multirow}
\usepackage{listings}
\usepackage{algorithm}
\usepackage{float}

\newcommand{\xmark}{\ding{55}}%
\newcommand{\cmark}{\ding{51}}%

\title{The Algonauts Project 2023 Challenge: UARK-UAlbany Team Solution}

\date{} 					

\author{Xuan Bac Nguyen\\
	CVIU Lab, EECS Department\\
	University of Arkansas, USA\\
	\texttt{xnguyen@uark.edu} \\
        \And
	{Xudong Liu} \\
 \\
	\texttt{cosinexd@gmail.com} \\
	\And
	{Xin Li} \\
	University at Albany, SUNY \\
	\texttt{xli48@albany.edu} \\
        \And
	{Khoa Luu} \\
	CVIU Lab, EECS Department\\
	University of Arkansas, USA\\
	\texttt{khoaluu@uark.edu} \\
}

\date{}

\renewcommand{\headeright}{Technical Report}
\renewcommand{\undertitle}{Technical Report}
\renewcommand{\shorttitle}{UARK-UAlbany Team}

\begin{document}
\maketitle

\begin{abstract}
This work presents our solutions to the Algonauts Project 2023 Challenge. The primary objective of the challenge revolves around employing computational models to anticipate brain responses captured during participants' observation of intricate natural visual scenes. The goal is to predict brain responses across the entire visual brain, as it is the region where the most reliable responses to images have been observed. We constructed an image-based brain encoder through a two-step training process to tackle this challenge. Initially, we created a pretrained encoder using data from all subjects. Next, we proceeded to fine-tune individual subjects. Each step employed different training strategies, such as different loss functions and objectives, to introduce diversity. Ultimately, our solution constitutes an ensemble of multiple unique encoders. The code is available at \url{https://github.com/uark-cviu/Algonauts2023}
\end{abstract}




\section{Introduction}
Over the past decade, the deep learning revolution has significantly impacted scientific research efforts, with profound implications for both artificial and biological intelligence. Initially inspired by the visual system of the mammalian brain, deep learning algorithms have evolved to become cutting-edge AI agents and scientific models for understanding the brain itself. As a result, research on artificial and biological intelligence is becoming increasingly intertwined. 

To improve this research direction, the 2023 edition of the Algonauts Project has been proposed \cite{gifford2023algonauts}. It follows the same objective as its predecessors in predicting human visual brain responses through computational models. However, it stands out from the previous 2021 edition of challenges \cite{cichy2021algonauts} using the Natural Scenes Dataset (NSD) \cite{allen2022massive}, the most extensive and data-rich collection of neural responses to natural scenes. The primary focus of the challenge lies in visual scene understanding, as vision remains an unsolved problem in both artificial and biological intelligence. The collaboration between these two fields has been particularly impactful in this domain, making it a promising area for further exploration.

\section{Dataset and Evaluation Protocol}. 
\subsection{Dataset}
The competition exploits the NSD dataset, an extensive collection comprising responses from 8 subjects, recorded using high-quality 7T fMRI. At the same time, they were exposed to approximately 73,000 different natural scenes to build the encoding models for the visual brain. From subjects 1 to 8, each has 9841, 9841, 9082, 8779, 9841, 9082, 9841, and 8779 unique images, respectively. In addition, the corresponding fMRI visual responses of each image are also provided. These signals contain the left hemisphere (LH) and right hemisphere (RH) that consist of 19,004 and 20,544 vertices, respectively, except for subject 6, which has 18,978 LH vertices and 20,220 RH vertices, and subject 8, which has 18,981 LH vertices and 20,530 RH vertices. These variations in the number of vertices are due to missing data for the specified subjects.

\subsection{Evaluation Metric} 
The evaluation metric is measured by taking the mean noise-normalized encoding accuracy across all the vertices of all subjects and hemispheres.
\begin{equation}
    m = \frac{1}{v} \sum_i^v \frac{R_i^2}{NC_i}
    \label{eq:metric}
\end{equation}
where $R_i$ is the Pearson correlation coefficient between predicted response $P_i$ and ground truth $G_i$ and $NC_i$ is the noise ceiling.

\section{Methods}
In this section, we first present the details of the pre-training stage using the data from all subjects. The fine-tuning stage will be discussed in the next section.

\subsection{All Subject Pretraining}

\begin{figure}[ht]
    \centering
    \includegraphics[width=0.95\textwidth]{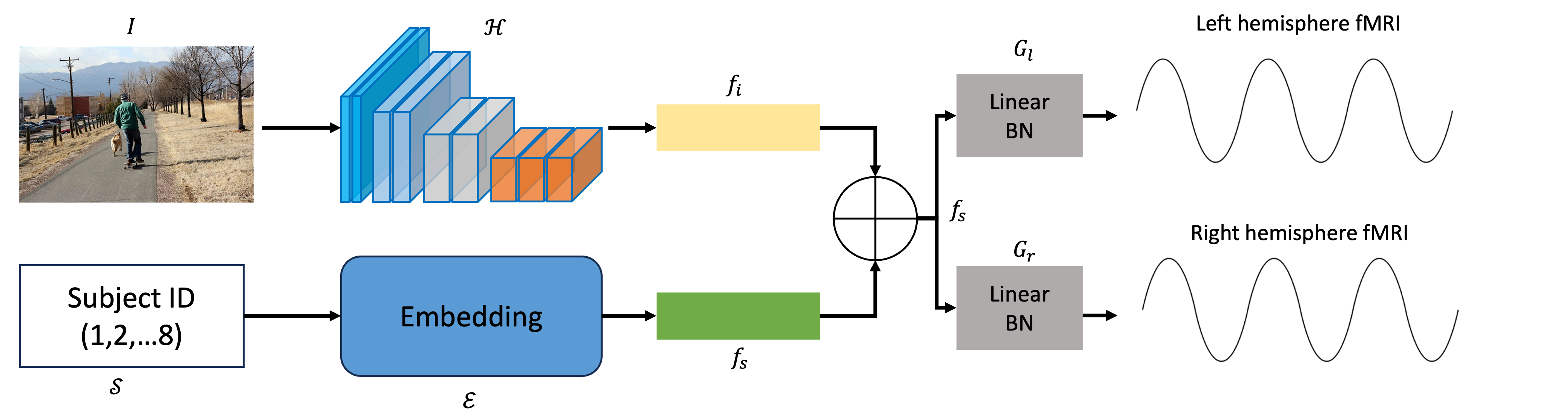}
    \caption{An overview of pretaining stage.}
    \label{fig:pretraining}
\end{figure}

The overview of the pretraining stage is illustrated in Fig \ref{fig:pretraining}.

Let $I \in \mathbb{R}^{H \times W \times C}$ be the input image where $H, W$, and $C$ are the height, width, and number of channels. We use a deep neural network (DNN) denoted $\mathcal{H}$ to extract the features of this image denoted $f_i$. 
\begin{align}
    f_i = \mathcal{H}(I) \in \mathbb{R}^{d_i}
\end{align}
In our experiment, a unique set of 1000 images was shared among eight subjects. However, training these images during the pre-training stage might encounter noise-related problems, as the brain responses to the same input image can differ between subjects. In order to effectively tackle this issue, the proposed network needs to incorporate the subject's ID as a crucial factor in learning and specify the corresponding fMRI signals accurately. Let $\mathcal{S}$ be the subject ID whose value ranges from 0 to 7, indicating the subject from 1 to 8. We design an embedding module $\mathcal{E}$ to learn the features of the subject $f_s$. 
\begin{align}
    f_s = \mathcal{E}(S) \in \mathbb{R}^{d_s}
\end{align}
Next, we concatenate the image and subject features to construct the features $f$. 
\begin{align}
    f = concat(f_i, f_s) \in \mathbb{R}^{d_i + d_s}
\end{align}
These features will be passed into two dependent blocks of linear and batch norm layers denoted as $G_l$ and $G_r$.
\begin{align}
    y_l = G_l(f) \in \mathbb{R}^{L_l} \\
    y_r = G_r(f) \in \mathbb{R}^{L_r}
\end{align}
where $y_l$ and $y_r$ are the predicted fMRI signals of the left and right hemispheres, respectively. 

\subsection{Fine-tuning on Individual Subject}
\begin{figure}[ht]
    \centering
    \includegraphics[width=0.95\textwidth]{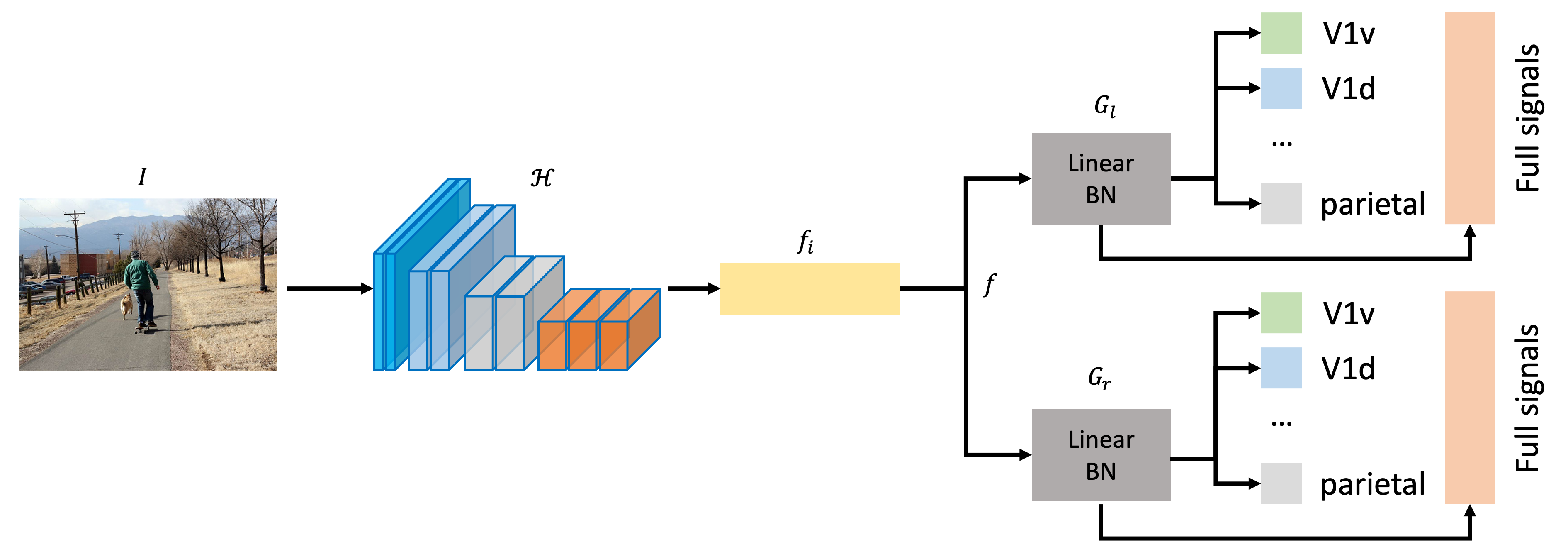}
    \caption{An overview of the fine-tuning stage.}
    \label{fig:finetuning}
\end{figure}
The overview of the fine-tuning stage is illustrated in Fig \ref{fig:finetuning}. At this stage, we used the pre-trained weight of DNN $\mathcal{H}$ and fine-tuned the individual data. Apart from the previous stage and predicting the full fMRI signals, we also add multiple heads, i.e., fully connected layers, to predict each signal of each region-of-interests (ROIs), e.g., V1v, V1d, etc. The final prediction is an average of each individual vertices and the full signals. 

\subsection{Loss function}
We utilize several loss functions in both stages. Along with the Smooth L1 loss function, we also implement the new loss functions as follows. 

\textbf{ Mean Normalized Pearson Correlation (MNNPC) Loss}. We implement the MNNPC loss function based on the evaluation metric defined in Eq. \eqref{eq:metric}. The detailed implementation in PyTorch is described as in Algorithm \eqref{algo:loss}.

\begin{algorithm}[!ht]
   \caption{Mean Noise-Normalized Pearson Correlation Loss}
   \label{algo:loss}
    \definecolor{codeblue}{rgb}{0.25,0.5,0.5}
    \lstset{
      basicstyle=\fontsize{7.2pt}{7.2pt}\ttfamily\bfseries,
      commentstyle=\fontsize{7.2pt}{7.2pt}\color{codeblue},
      keywordstyle=\fontsize{7.2pt}{7.2pt},
    }
\begin{lstlisting}[language=python]
class MNNPC(nn.Module):
    def __init__(self):
        super().__init__()

    def forward(self, pred, gt, nc=None):
        pred_mean = torch.mean(pred, axis=0)
        gt_mean = torch.mean(gt, axis=0)

        gt_t = gt.T
        pred_t = pred.T

        ts = (gt_t - gt_mean.view(-1, 1)) * (pred_t - pred_mean.view(-1, 1))
        ts = ts.sum(axis=1)

        ms1 = (gt_t - gt_mean.view(-1, 1)) ** 2
        ms1 = ms1.sum(axis=1)

        ms2 = (pred_t - pred_mean.view(-1, 1)) ** 2
        ms2 = ms2.sum(axis=1)
        ms = (ms1 * ms2) ** 0.5

        rv = ts / (ms + 1e-8)
        rv = 1 - (rv + 1) / 2

        if nc is not None:
            nc[nc == 0] = 1
            rv = rv**2 / torch.from_numpy(nc).to(rv.device)

        return rv.mean()
\end{lstlisting}
\end{algorithm}
\textbf{Pearson Correlation Loss}. We also implement the Pearson Correlation Loss which aims to optimize the challenge metric indirectly. The details of the implementation are shown in algorithm \ref{algo:pcc_loss}

\begin{algorithm}[!ht]
   \caption{Pearson Correlation Loss}
   \label{algo:pcc_loss}
    \definecolor{codeblue}{rgb}{0.25,0.5,0.5}
    \lstset{
      basicstyle=\fontsize{7.2pt}{7.2pt}\ttfamily\bfseries,
      commentstyle=\fontsize{7.2pt}{7.2pt}\color{codeblue},
      keywordstyle=\fontsize{7.2pt}{7.2pt},
    }
\begin{lstlisting}[language=python]
class PCLoss(nn.Module):
    def __init__(self):
        super().__init__()
        self.cos = nn.CosineSimilarity(dim=1, eps=1e-6)

    def forward(self, pred, gt, nc=None):
        pearson = self.cos(
            x1 - x1.mean(dim=1, keepdim=True), x2 - x2.mean(dim=1, keepdim=True)
        )
        pearson = pearson.mean()
        pearson = (pearson + 1) / 2
        return 1 - pearson
\end{lstlisting}
\end{algorithm}

\section{Experiments and Results}
\subsection{Implementation Details}
For each subject, we split the database into 5 folds where there are approximately 3900 and 1900 samples in training and validation. In both the pre-training and fine-tuning stages, the images are resized to $384 \times 384$. We select the embedding vector dimension of the subject $d_s = 512$ while the image feature dimension $d_i$ depends on the DNN $\mathcal{H}$. All the code is easily implemented in PyTorch framework and trained by an A100 GPU. The learning rate is initially set to $0.0001$ and then gradually reduced to zero under the ConsineLinear \cite{cosinelr} policy. The batch size is set to $8$/GPU. The model is optimized by AdamW \cite{adamw} for $12$ epochs or until the network meets the early stopping. The pertaining and fine-tuning are completed within 8 hours for each subject approximately.

\subsection{Experiment Results}
We have chosen \texttt{convnext\_xlarge} \cite{liu2022convnet} as the baseline for our experiment and \texttt{nn.Embedding} as the embedding $\mathcal{E}$. The results are presented in Table \ref{tab:result}. The baseline, which is based solely on the Smooth L1 loss function without any fine-tuning, achieves a submission score of 54.21\%. However, by implementing a pre-training strategy, we observe an improvement of approximately 2.3\%, bringing the score to 56.5\%. Further enhancements are achieved by introducing two new loss functions, PCLoss and MNPCLoss, resulting in a submission score of 57.01\%.
\subsection{Ensemble}
\begin{figure*}[ht]
    \centering
    \includegraphics[width=1.0\textwidth]{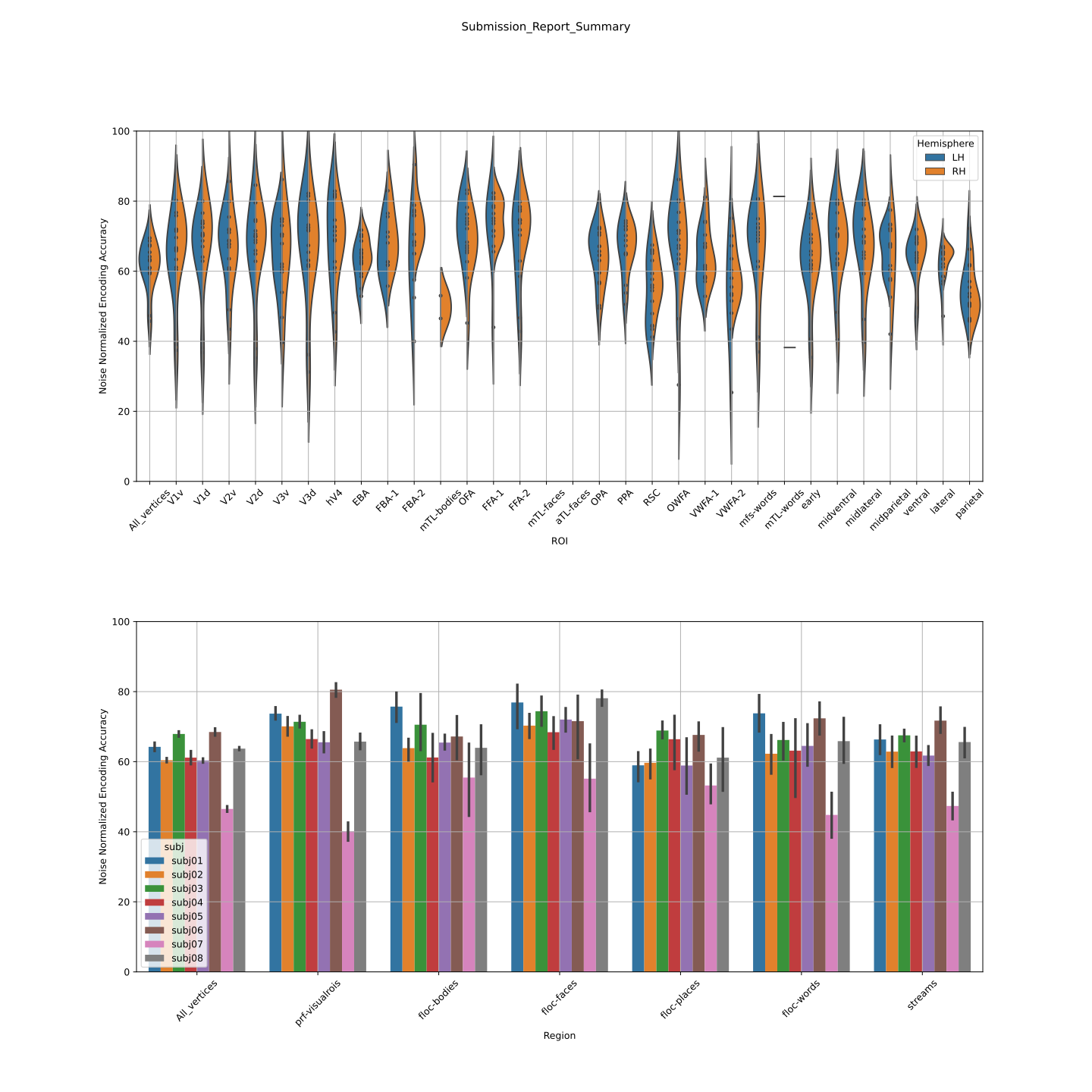}
    \caption{Submission report summary.}
    \label{fig:submission_summary}
\end{figure*}

In addition to training \texttt{convnext\_xlarge} as the baseline, we experimented with several other backbones, including \texttt{seresnext101d\_32x8d} \cite{hu2018squeeze}, \texttt{convnext\_base}\cite{liu2022convnet}, \texttt{vit\_small\_patch16\_224} \cite{dosovitskiy2020image}, \texttt{efficientnet\_b7} \cite{tan2019efficientnet}, \texttt{seresnet152d}\cite{hu2018squeeze} and \texttt{seresnextaa101d\_32x8d}\cite{hu2018squeeze}. Each backbone underwent various training settings, including different combinations of loss functions, to further increase the diversity of the ensemble. Finally, a heuristic approach is used to compute the weighted average of all models. Consequently, the final submission achieves a submission score of 61.56\%. Details of the submission results are illustrated in Fig. \ref{fig:submission_summary}

\begin{table}[!ht]
\centering
\begin{tabular}{|l|l|l|l|l|l|}
\hline
Backbone $\mathcal{H}$      & Pretraining & L1 Loss  & PCLoss & MNPCLoss & Score (\%) \\ \hline
\texttt{convnext\_xlarge}  & \xmark      & \cmark & \xmark     & \xmark & 54.21            \\ \hline
\texttt{convnext\_xlarge}  & \cmark         & \cmark & \xmark     & \xmark       & 56.50            \\ \hline
\texttt{convnext\_xlarge}  & \cmark         & \cmark & \cmark    & \cmark      & 57.01            \\ \hline
\end{tabular}
\label{tab:result}
\caption{Summary of experiment results on various loss functions and pretraining strategy}
\end{table}

\section{Conclusion and Discussion}

Our solution makes significant contributions to the challenge in several key aspects. First, we introduce a pretraining stage that greatly enhances performance. Additionally, we propose two novel loss functions that effectively optimize the evaluation metrics. During the competition, we also identify several intra-challenges that need to be addressed. Notably, we observe that the choice of the backbone has a substantial impact on the final predictions. To improve upon our solution, future work will be delved into exploring network designs specifically tailored to the problem. Furthermore, we are interested in investigating the utilization of additional information from raw data for self-supervised learning, which could open promising research directions.

\section*{Acknowledgments}
Our sincere gratitude goes to the Arkansas High-Performance Computing Center for generously providing GPUs for this challenge.


\bibliography{library}

\bibliographystyle{abbrv}

\end{document}